\documentclass[10pt,twocolumn,letterpaper]{article}

\usepackage{authblk}
\usepackage{multirow}
\usepackage[accsupp]{axessibility}  
\usepackage[pagenumbers]{iccv} 

\definecolor{iccvblue}{rgb}{0.21,0.49,0.74}
\usepackage[pagebackref,breaklinks,colorlinks,allcolors=iccvblue]{hyperref}

\def\paperID{CVAUI \& AAMVEM-22} 
\def\confName{ICCV}
\def\confYear{2025}

\title{KAMERA: Enhancing Aerial Surveys of Ice-associated Seals\\in Arctic Environments}

\author[1]{Adam Romlein}
\author[2]{Benjamin X. Hou}
\author[3]{Yuval Boss}
\author[3]{Cynthia L. Christman}
\author[2]{Stacie Koslovsky}
\author[2]{Erin E. Moreland}
\author[1]{Jason Parham}
\author[1]{Anthony Hoogs}

\affil{Kitware, Inc., USA, \{adam.romlein,jason.parham,anthony.hoogs\}@kitware.com}
\affil[2]{NOAA NMFS AFSC MML, USA, \{ben.hou,stacie.koslovky,erin.moreland\}@noaa.gov}
\affil[3]{CICOES, University of Washington, USA, yuval@uw.edu, cynthia.christman@noaa.gov}

\begin{document}
\maketitle
\begin{abstract}
We introduce KAMERA: a comprehensive system for multi-camera, multi-spectral synchronization and real-time detection of seals and polar bears. Utilized in aerial surveys for ice-associated seals in the Bering, Chukchi, and Beaufort seas around Alaska, KAMERA provides up to an 80\% reduction in dataset processing time over previous methods. Our rigorous calibration and hardware synchronization enable using multiple spectra for object detection. All collected data are annotated with metadata so they can be easily referenced later. All imagery and animal detections from a survey are mapped onto a world plane for accurate surveyed area estimates and quick assessment of survey results. We hope KAMERA will inspire other mapping and detection efforts in the scientific community, with all software, models, and schematics fully open-sourced.

\end{abstract}    
\section{Introduction}
\label{sec:intro}

Crewed and Uncrewed Aerial Systems (UAS) have emerged as popular methods for data collection, as they allow for sampling across large geographic areas that are inaccessible or impractical to traverse via other platforms (e.g., boats, automobiles). Aerial survey methods are commonly employed in agriculture, civil engineering, geology, and the biological sciences \cite{del2021unmanned, hoskere2019vision, siebert2014mobile, tziavou2018unmanned, cole2007methodologies, madore2018noaa, davila2022adaptopensourcesuaspayload, hann2024noaa}. Increasingly, such surveys rely upon the integration of robust onboard computing systems paired with remote sensing instrumentation (e.g., LIDAR, cameras, radar) to capture high-quality, rich datasets that can be leveraged to streamline data analyses and information extraction, thereby reducing the time between data collection and decision-making. For these reasons, many wildlife management agencies have shifted towards use of such technologies as they enable more timely interventions. In particular, multi-spectral aerial surveys with AI methods for object detection assistance have become a common approach for studies of abundance and distribution of ice-associated seals \cite{eikelboom2019improving, parham2017animal, rubensteinGreatZebraGiraffe2015, berger2016great, crall2013hotspotter, morrison2016individual, parhamAnimalDetectionPipeline2018, dawkinsAutomaticScallopDetection2013}. Their broad distribution across expansive sea ice habitat presents significant practical challenges for data collection and expedient analysis of multi-terabyte imagery datasets.

\begin{figure}[t]
  \centering
   \includegraphics[width=1\linewidth]{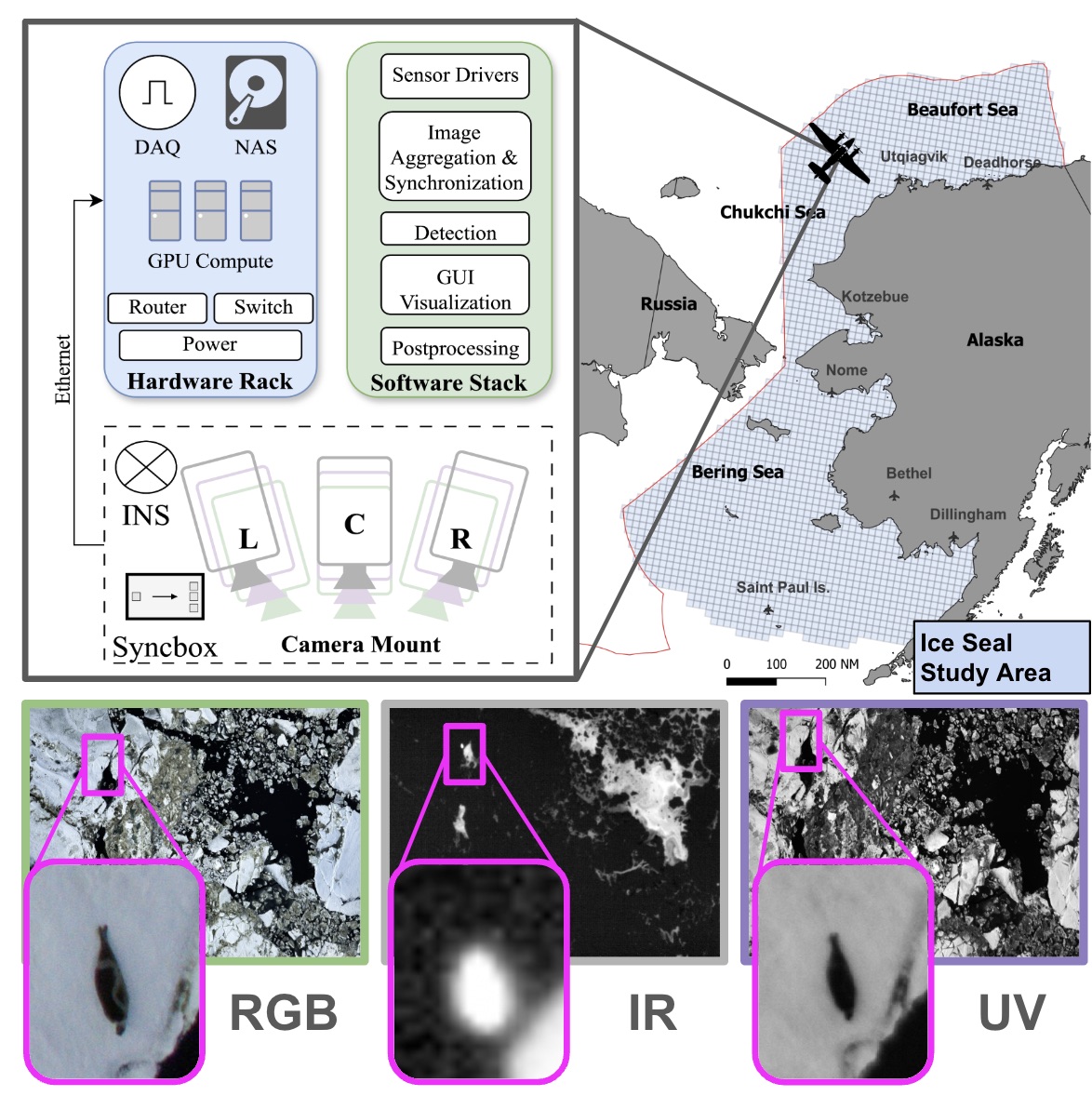}
   \caption{KAMERA's hardware, software, and area of operations outlined. A ribbon seal captured with KAMERA is shown in RGB, IR, and UV on the bottom. This aerial survey used three cameras in each modality to capture a wide swath across the flight path at high resolution.}
   \label{fig:concept}
\end{figure}

\begin{table*}
  \centering
  \small
  \begin{tabular*}{\linewidth}{@{\extracolsep{\fill}} @{}lllllll@{} } 
    \toprule
    Survey Region & Year (s) & RGB Cameras & Collection & Data Sync & Processing & \begin{tabular}[l]{@{}l@{}}Time to\\Results\end{tabular}\\
    \midrule
    \textbf{Bering Sea} & 2012-2013 & Digital SLRs & 1.8 million images & Independent & \begin{tabular}[l]{@{}l@{}}IR Peak Analysis,\\Blob Detection\end{tabular} & 1-2 yrs \\
    \midrule
    \textbf{Chukchi Sea} & 2016 & \begin{tabular}[l]{@{}l@{}}Machine Vision\\(29MP)\end{tabular} & 1 million pairs & Integrated & \begin{tabular}[l]{@{}l@{}}Semi-Automated\\Hot Spot Detection\end{tabular} & 6 mos \\
    \midrule
    \textbf{Beaufort Sea} & 2021 & \begin{tabular}[l]{@{}l@{}}Machine Vision\\(29MP)\end{tabular} & 900,000 triplets & \textbf{KAMERA} & \begin{tabular}[l]{@{}l@{}}YOLOv3 IR/RGB\\Trigger Model\end{tabular}  & 5 wks\\
    \midrule
    \begin{tabular}[l]{@{}l@{}}\textbf{Bering, Chukchi, and}\\\textbf{Beaufort Seas}\end{tabular} & 2025 & \begin{tabular}[l]{@{}l@{}}Medium Format\\(120MP)\end{tabular} & \begin{tabular}[l]{@{}l@{}}1.5 million samples, \\reduced\end{tabular} & \textbf{KAMERA} & \begin{tabular}[l]{@{}l@{}}YOLOv3 IR\\Hot Spot Detection\end{tabular} & TBD\\
    \bottomrule
  \end{tabular*}
  \caption{Timeline of large-scale, image-based surveys for ice-associated seals in U.S. Arctic and subarctic waters off Alaska.}
  \label{fig:timeline}
\end{table*}

Our paper describes the design of a novel multi-camera, multi-spectral aerial imaging system capable of capturing, archiving, and performing real-time object detection and classification called KAMERA: the  \underline{K}nowledge-guided Image \underline{A}cquisition \underline{M}anag\underline{ER} and \underline{A}rchiver, see Figure \ref{fig:concept}. Previous systems used for surveys were limited by their loose coupling of proprietary software and hardware systems, resulting in asynchronous collection and poor data alignment impeding information extraction by automated methods. Additionally, their closed-source, proprietary nature limits broader utility to the scientific community. KAMERA improves upon these methods in the following ways:

\begin{itemize}
  \item \textbf{Multi-Camera, Multi-Spectral Synchronization} All data is collected under a single external time pulse and aggregated into one storage location, meticulously labeled with necessary metadata.
  \item \textbf{Real-time Detection} Onboard GPUs are used to analyze this synchronized imagery enabling real-time decisions on which data to archive.
   \item \textbf{Mapping} All imagery and detections are mapped for accurate survey area calculation and post-flight data evaluation.
  \item \textbf{Open-Source} All software has been open-sourced under the Apache License (Version 2.0) and pulls together numerous different off-the-shelf camera drivers and hardware specifications.
\end{itemize}

In this paper, we begin by providing a brief background on aerial surveys for ice-associated seals and the motivation for the development of KAMERA. We then detail the development, components, and capabilities of our system, including a discussion of the AI model development and implementation. Finally, we conclude with the real-world results of our models and the future of KAMERA.

\section{Background}
\label{sec:background}
Ice-associated seals (ringed, ribbon, spotted, and bearded seals) are broadly distributed across an expansive sea ice habitat of Arctic and subarctic regions. These species are important resources for Arctic Indigenous communities and play an important role in the ecosystem \cite{ipcomm2006}. With projected changes to their sea ice habitat, two of these species (ringed and bearded) have been listed as Threatened under the Endangered Species Act\footnote{\url{https://www.law.cornell.edu/uscode/text/16/chapter-35}} and all marine mammals are protected under the Marine Mammal Protection Act \cite{mmpa1972}. Monitoring these species in U.S. waters is therefore required by law to inform population management. 

On average, an ice seal aerial survey collects around a million samples (RGB/IR pairs or RGB/IR/UV triplets) covering approximately 20,000 km of survey area. However, less than 1\% (or 10,000) of these images typically contain seals, necessitating the use of methods that allow for semi- or fully-automated processing of image sets. Towards this end,  early surveys for ice-associated seals (see Table \ref{fig:timeline}) employed thermal cameras capturing video,  digital SLR cameras capturing images at one hertz, and a GPS logging position every five seconds \cite{moreland2013bering, doi:10.1139/as-2024-0068, sigler2015advances}. While  effective, these systems required considerable manual effort for data processing and information extraction.


Development on KAMERA began in 2018, with the mission of designing a comprehensive multi-camera, multi-spectral synchronized data collection system. The first successful survey with KAMERA was in 2021, where around 900,000 samples (image triplets) were collected over the Southern Beaufort Sea. We utilized a two-stage detection pipeline that first identifies thermal hot spots in the infrared (IR) imagery, crops out the associated portion of the paired color image, and runs an additional model to identify and classify the species of the seal. The integration of AI methods with the improved data produced by KAMERA enabled an 80\% reduction in time-to-results compared to the Chukchi Sea surveys in 2016 of similar size.

Between 2021 and 2025, KAMERA underwent several upgrades in both software and hardware with numerous testing and evaluation steps. The color cameras were upgraded from Prosilica GT6600 machine vision cameras to Phase One iXM-GS120 medium format cameras with finer ground sample distance (GSD), improving species classification. The compute systems were modernized to support the processing of this new, larger imagery. In addition, we made significant software and GUI updates to improve robustness, reproducibility, and usability of the system.

The survey conducted in 2025 of all three Alaskan seas was the largest concurrent survey of the area ever conducted. Two aircraft flew from the cities of Nome, Bethel, Kotzebue, Barrow, and Deadhorse, Alaska, collecting 1.5 million samples over the course of the survey. One aircraft held a 9-camera system capturing image triplets, and the other held a 6-camera system capturing image pairs. We opted to use the IR hot spot detection model in lieu of the two-stage IR-RGB model due to our color camera instrumentation upgrade. The presence of high-scoring thermal objects formed the basis for image archiving, reducing the total data volume.  The data collected from this survey will be incorporated into training a new IR-RGB model. Data extracted from the imagery along with processed survey data are publicly available at \cite{noaa2024polar}. 
\section{System development}
\label{sec:project}

The KAMERA system includes the hardware we used to accomplish our particular task of ice-associated seal surveys. While the hardware we selected is dependent on our goal, the software is designed to be broadly applicable, with emphasis on providing precise, synchronized image collection and offering interoperability of detection models. In analysis of the data, the object detector can always be upgraded, exchanged, or detections annotated manually after the fact.

\begin{figure}[t]
  \centering
   \includegraphics[width=1\linewidth]{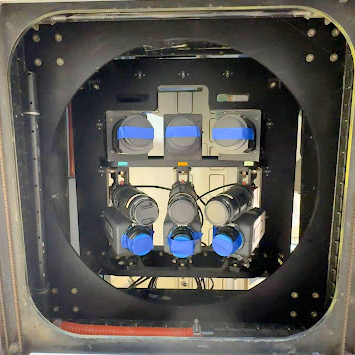}
   \caption{Camera mount containing the nine-camera sensor system for KAMERA's crewed system installed in the belly of a King Air aircraft.}
   \label{fig:cameramount}
\end{figure}

\subsection{System hardware \& software}

Ice seals reliably produce strong heat signatures against the chilly backdrop of Arctic sea ice, but the low resolution and lack of color in thermal cameras preclude reliable species classification. Color cameras have significantly higher resolution and capture visible features, but it is difficult to find seals via either automated or manual means, due to the scale and complexity of the imagery. Polar bears and white-coat lanugo seal pups are challenging to detect in color and thermal imagery due to their white fur and low thermal signature, but can be distinguished in the ultraviolet spectra as their fur absorbs UV light. All configurations presented are from the nine-camera system onboard a King Air crewed aircraft, but other, more minimal, configurations exist for a Twin Otter crewed aircraft and a NASA SIERRA-B UAV \cite{borade2021nasa}. All hardware is summarized in Table ~\ref{tab:parts}.

\begin{table}
  \centering
  \begin{tabular}{@{}lc@{}c@{}rc@{}}
    \toprule
    Part & Model & Cost\\
    \midrule
    Color Camera & Phase One iXM-GS120 & 3 x \$60k \\
    Color Lens & Schneider-Kreuznach RS-110mm & 3 x \$11k \\
    Thermal Camera & FLIR A6751 SLS & 3 x \$115k \\
    UV Camera & Prosilica GT4907 & 3 x \$9k \\
    UV Lens & Jenoptik 105mm UV-VIS & 3 x \$6k \\
    GPU Compute & Nuvo-10208GC & 3 x \$7k \\
    NAS & Synology FS 1018 & 1 x \$5k \\
    INS & Applanix POS AVX 210 & 1 x \$20k \\
    DAQ & MCC USB-2408-2AO & 1 x \$1k \\
    Router & Mikrotik hEX RB750Gr3 & 1 x \$50 \\
    Switch & Trendnet TEG-S591 & 1 x \$100 \\
    Syncbox & Custom & 1 x \$200 \\
    \midrule
    Total & & \$650k \\
    \bottomrule
  \end{tabular}
  \caption{KAMERA's crewed system components.}
  \label{tab:parts}
\end{table}

\begin{figure*}
  \centering
  \begin{subfigure}{0.48\linewidth}
    \includegraphics[width=1\linewidth]{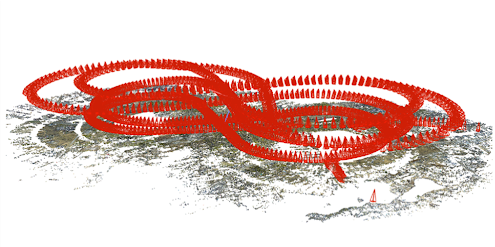}
    \label{fig:colmap-a}
  \end{subfigure}
  \begin{subfigure}{0.48\linewidth}
    \includegraphics[width=1\linewidth]{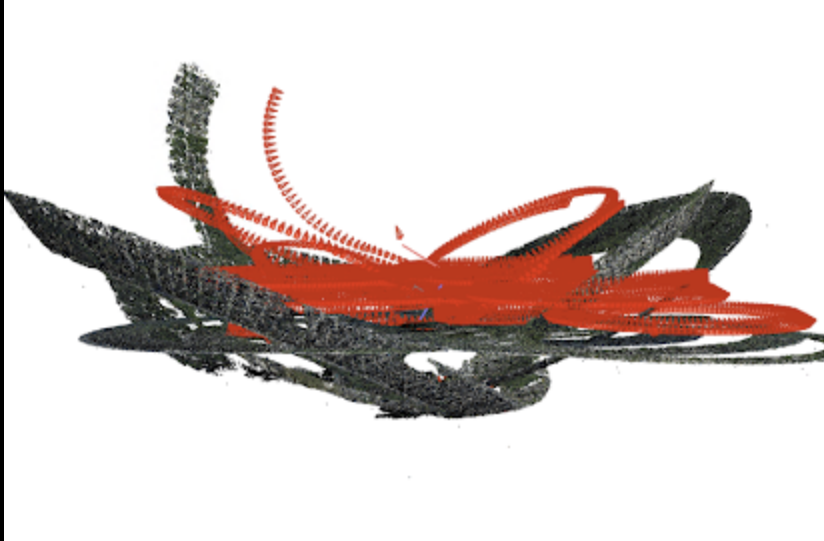}
    \label{fig:colmap-b}
  \end{subfigure}
  \caption{Analysis of the imagery on a calibration flight over Anchorage, AK using COLMAP. (Left) Example of a successful COLMAP sparse reconstruction, with large portions of the figure eight reconstructed onto a planar surface. (Right) Example of a failed COLMAP sparse reconstruction, with non-planar contours.}
  \label{fig:colmap}
\end{figure*}

\textbf{Cameras}
The Phase One iXM-GS120 color camera provides the GSD of 1 - 1.7 cm/px (range from center nadir to edge of angled cameras), which is sufficient for animal identification and species classification at survey altitude (approx. 1,000ft). The features of its global shutter, shutter speed, and high dynamic range are ideal for aerial imagery structure-from-motion reconstructions and animal detection. The FLIR-A6751 produces thermal imagery at a high enough resolution to spot seal pups, with an integration time low enough to avoid excessive blur. The Prosilica GT4907 UV camera has the shutter speed and resolution required to capture polar bears and the white-coat seal pups. All three of these cameras can be triggered externally, fundamental for the synchronization required. We decided on three cameras of each spectrum (nine in total) both to maximize the area covered in a single aerial transect and to provide redundancy in case of hardware failure.

\textbf{Sensors}
An Inertial Navigation System (INS) is essential for geolocating the animal detections in the imagery. The Applanix POS AVX 210 produces a 12-degrees-of-freedom output suitable for accurate projection and allows for an external trigger to be timestamped with the current GPS time. This enables an association between images, trigger, and absolute GPS UTC time. A custom synchronization box was created to split the pulse trigger from the Data AcQuisition device (DAQ) to all nine cameras and the INS.

\textbf{Compute \& Software}
Three ruggedized compute systems serve as the backbone of the processing, each handling a single triplet of cameras (RGB, IR, and UV). The Neousys Nuvo 10208GC offered space for up to two GPUs for future expansion in a vibration and shock MIL-STD rated chassis. The sensor data is directly written to internal NVMe SSDs, and incrementally copied over to an onboard Synology Networked Attached Storage (NAS) equipped with SSD drives to allow real-time write speed. An onboard MikroTik router and TRENDnet 2.5Gb switch provide the connectivity needed between all devices. We use the Robot Operating System (ROS) \cite{doi:10.1126/scirobotics.abm6074} as our backbone middleware, given its broad community support and flexibility in integration. All camera drivers are written in C++, and data aggregation processes in Python. Ansible and Docker \cite{merkel2014docker} were chosen to create reproducible systems from a base OS. We utilize a software stack of Supervisor, Bash, tmux, and Docker Compose for system startup and control. 

\textbf{Aircraft integration}
The compute, DAQ, NAS, router, network switch, and DC and AC power distribution for each system are affixed in racks to the floor of the aircraft. The cameras, INS, and DC power for both are rigidly affixed to a camera mount that installs into the belly of the aircraft as shown in Figure \ref{fig:cameramount}. The  camera mounts have physical stops that allow for different camera angles. This capability allows for varied side-to-side overlap to suit the needs of different research projects. With this camera configuration, we must be able to associate pixels from one spectrum to another for multi-spectral detection. This requires knowing the positions of each camera relative to another, and culminates in a multi-camera calibration.

\subsection{Calibration}

The goal of our calibration process is to obtain the intrinsic parameters of each of the nine cameras (focal length, focal point, distortion) and to estimate a rigid transformation between the INS and each camera. With these two pieces of information, at any instance of time with a synchronized pair of INS pose and imagery, any pixel can be projected out onto a planar model of the world. We utilize \textbf{COLMAP} \cite{schoenberger2016sfm,schoenberger2016mvs} to estimate both pieces during its structure-from-motion (SfM) sparse reconstruction process, which uses \textbf{SIFT} \cite{sift2004} features. SIFT is limited in areas of sparse features and between spectrums, so we are considering deep-learning-based features and keypoints such as SuperPoint \cite{detone2018superpointselfsupervisedpointdetection}, SuperGlue \cite{sarlin2020supergluelearningfeaturematching}, and ALIKED \cite{zhao2023alikedlighterkeypointdescriptor} for future calibration efforts.

\textbf{Calibration Flight}
Initially a calibration flight is flown over a dense area, ideally a city due to a diversity of distinct features, delineated by three figure eights flown at 1,000, 2,000, and 3,000 ft increments. We collect the frames at around 50\% overlap for optimal image-to-image matching. If the reconstruction succeeds we can proceed to register the camera models generated to the INS. If the reconstruction fails, we must tune COLMAP matching parameters, or possibly revisit the data for a new collection. Examples of a successful and failed reconstruction are shown in Figure \ref{fig:colmap}. As long as the data has been collected in the specified figure eights at differing altitudes, the reconstruction is generally successful. Common failure cases included poor camera focus, not enough image overlap, and under / overexposed imagery.

\begin{figure}[t]
  \centering
      \begin{subfigure}{1\linewidth}
        \includegraphics[width=1\linewidth]{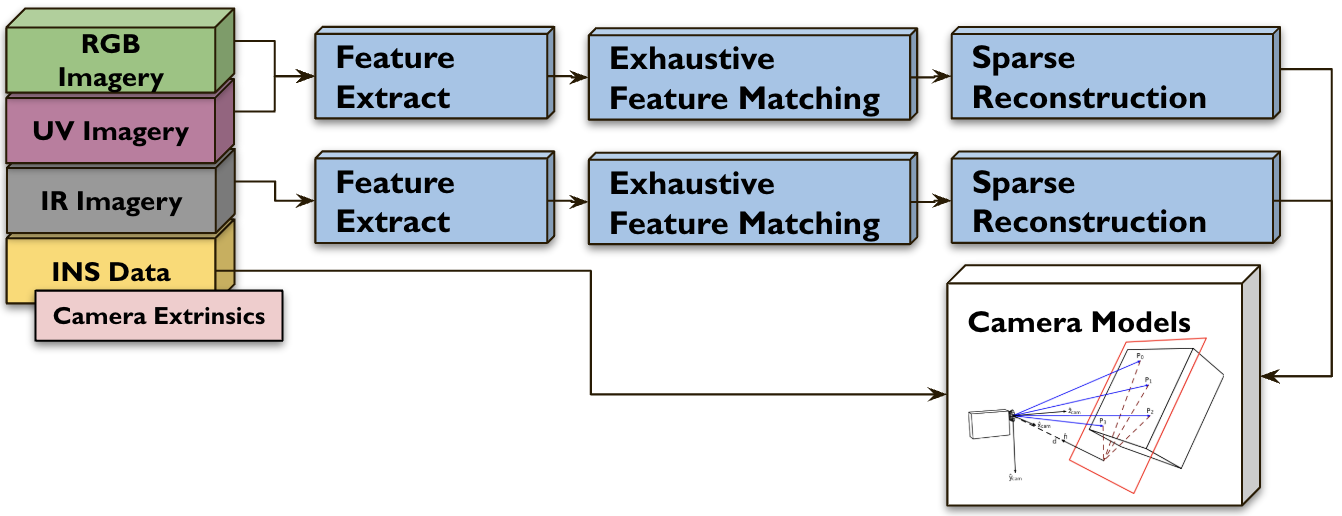}
        \label{fig:cal-a}
        \hrule
      \end{subfigure}
      \begin{subfigure}{1\linewidth}
        \includegraphics[width=1\linewidth]{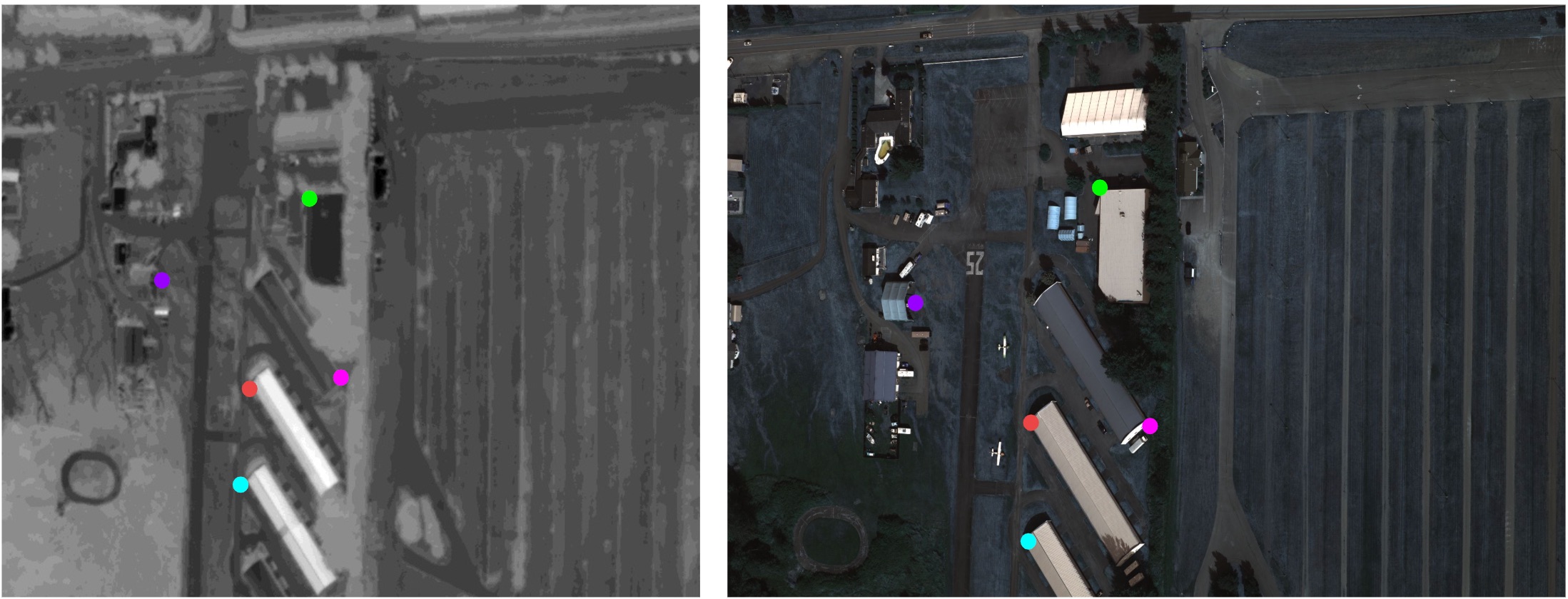}
        \label{fig:cal-b}
      \end{subfigure}
  \caption{(Top) Overview of the calibration process for the nine cameras, from feature extraction to camera model generation. (Bottom) Manual alignment for image pairs, with an IR image with selected features on the left, and matching RGB features in the color image on the right.}
  \label{fig:calibration}
\end{figure}

\textbf{Estimating Transforms}
An outline of the process to estimate transforms is shown in Figure \ref{fig:calibration} (top). SIFT features struggle to match between color and thermal, so we generate a standalone IR sparse reconstruction using just thermal imagery for IR calibration. RGB and UV SIFT features are generally compatible, so we combine these six cameras into a single sparse reconstruction. Once we have these reconstructions, they are only in relative world space, so we align each of these models to the real-world with COLMAP using the GPS positions we have captured with the INS at each timestep. Functionally, we have now built up a corpus of 2D-3D correspondences, and can estimate the rigid transformations between each camera and the INS required for accurate projection.

\textbf{Calibration Drift}
This calibration can drift over time due to multiple factors, including: unmounting / remounting cameras or lenses, adjusting camera angles, or any INS adjustments. We have two options to correct this. If it is a large change like swapping out a camera, we must repeat the whole calibration flight as above. If it is a small change, we can perform the manual alignment step outlined in Figure \ref{fig:calibration} (bottom) to ensure our relative alignment is exact, even if our absolute alignment may have drifted a small amount. In general, only one calibration is needed per survey. 

\textbf{Camera Models}
The final output of this process is nine YAML files containing the cameras' intrinsic and extrinsic parameters. These camera models are used to give us precise mapping results of square miles covered (in the form of footprints for each image), perform IR to RGB trigger-based detection, and geolocate animal detections.

\subsection{Detection of ice-associated seals \& polar bears}
\label{sec:detection}
Animal surveys in remote locations, especially over water and ice, are ideal candidates for object detection with deep learning. The vast majority of imagery will be empty, presenting an extraordinarily tedious human task of annotation that often results in fatigue and error. Conversely, the specific instances of animals stand out against the relatively homogeneous background. Operating in real time offers two main advantages: saved compute on the ground and the ability to drop imagery before it is even saved, eliminating large swathes of blank imagery that normally must be reviewed and catalogued. 

\newcommand{\anchors}{\mathcal{A}}
\begin{table}
  \centering
  \begin{tabular*}{\linewidth}{@{\extracolsep{\fill}} llll } 
    \toprule
    Spectrum & Classes & Model Layers & Input Size \\
    \midrule
    IR & Hot Spot & $P_3/8$ - 5$\anchors$ & 512×640×1  \\
    \midrule
    RGB & Polar Bear & \begin{tabular}[c]{@{}l@{}}$P_5/32$ - 3$\anchors$\\$P_4/16$ - 5$\anchors$\\$P_3/8$ - 1$\anchors$\end{tabular} & 416×416×3 \\
    \midrule
    RGB & \begin{tabular}[c]{@{}l@{}} Ringed Seal\\Bearded Seal\end{tabular} & \begin{tabular}[c]{@{}l@{}}$P_5/32$ - 3$\anchors$\\$P_4/16$ - 4$\anchors$\\$P_3/8$ - 3$\anchors$\end{tabular} & 512×512×3 \\
    \bottomrule
  \end{tabular*}
  \caption{Model architecture specifications for the three detection models. The IR hot spot model was trained with $\lambda_{noobj}$ = .5, and the RGB seal model was trained with Focal Loss. $\anchors$ = anchors. $P$ notation for different scale feature layers derived from \cite{lin2017featurepyramidnetworksobject}.}
  \label{fig:model_architecture}
\end{table}

Deep learning pipelines for automated detection of bearded seals, ringed seals, and polar bears were developed separate of the in-flight system and evaluated at regular intervals using \textbf{VIAME} \cite{dawkinsOpensourcePlatformUnderwater2017}, a do-it-yourself computer vision toolkit targeting marine species analytics. While we targeted marine mammals, any deep-learning detection model could be used through VIAME for other animal surveys, such as moose, elephants, or sea turtles.

\textbf{Dataset Formation}
Training data from prior surveys was manually labeled with a mix of bounding boxes and center points for use in model development and validation. An additional small test set was held out for final evaluation of the detection pipeline. For model development we had 31,000 ring seal annotations, 4,000 bearded seal annotations, and 300 polar bear annotations. Due to inconsistencies with image labeling and a large amount of missing labels, especially in the thermal imagery, multiple early models were ensembled and the resulting pseudolabels were used in training of the final models. The only annotations available were in IR and RGB, so no UV models were trained. UV models are currently under development for polar bears and white-coated seal pups.

\textbf{Model Development}
During model development in early 2019, we considered several real-time models, including  SSD \cite{Liu_2016}, FPN \cite{lin2017featurepyramidnetworksobject}, and RetinaNet-based detectors. We decided on the \textbf{YOLOv3} \cite{Redmon2018} architecture for its emphasis on real-time detection, and with some modifications, an architecture suitable for our task.

\begin{figure}[t]
  \centering
   \includegraphics[width=1\linewidth]{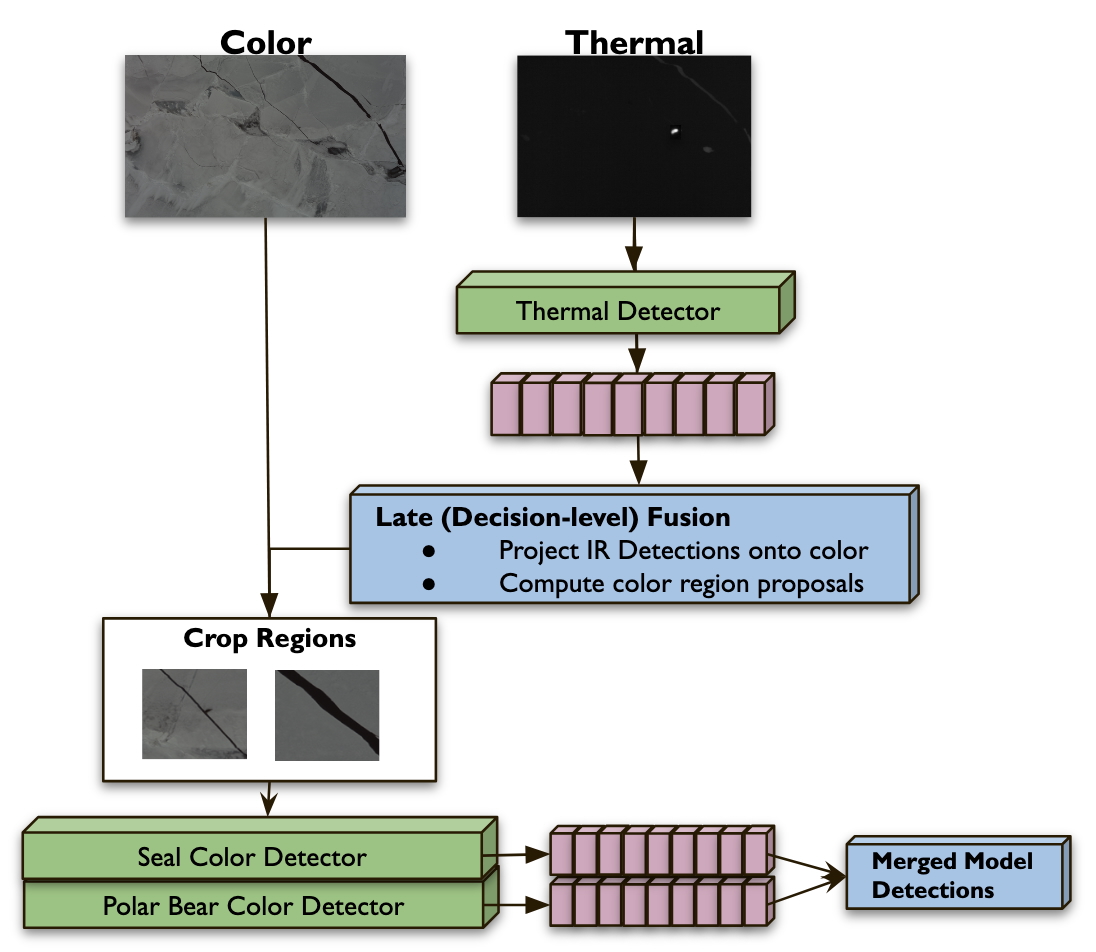}
   \caption{Late fusion 2-stage pipeline, utilizing thermal detections to crop out full-resolution chips of color imagery.}
   \label{fig:late_fusion}
\end{figure}

\begin{figure}[t]
  \centering
   \includegraphics[width=1\linewidth]{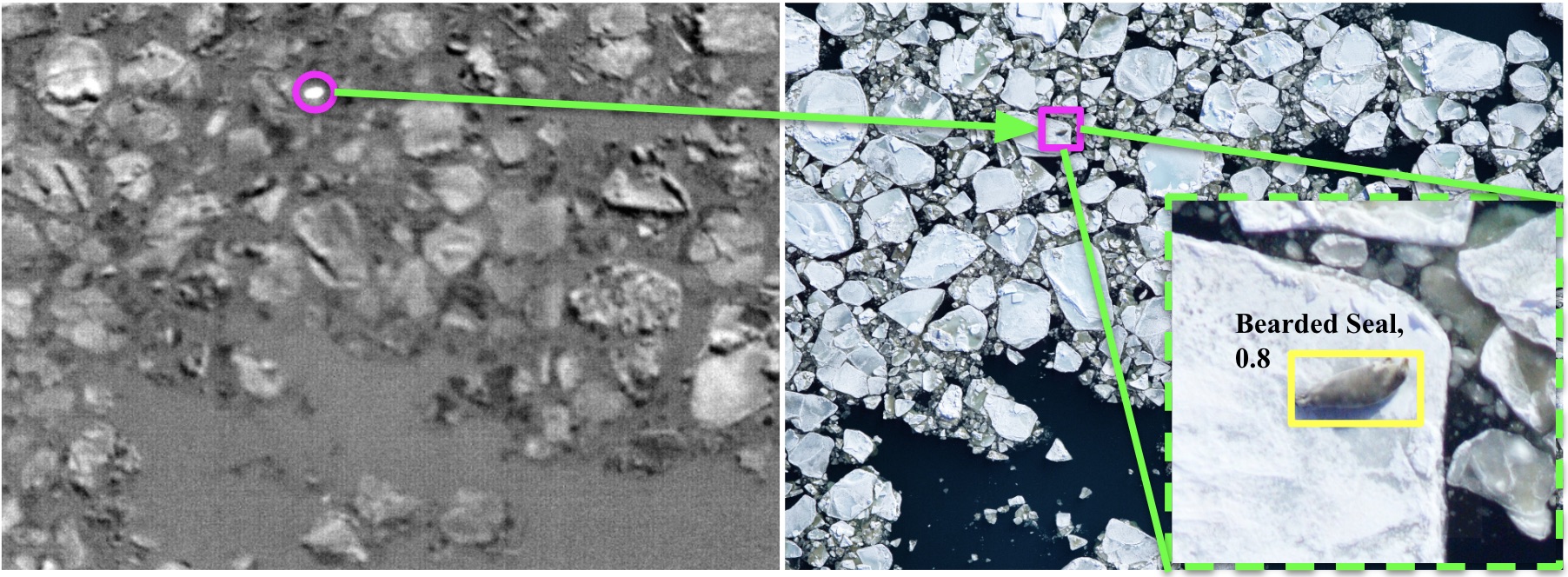}
   \caption{Visualization of late fusion pipeline. (Left) IR imagery with a thermal hot spot. (Right) Color imagery with the triggered crop from its higher resolution for species classification.}
   \label{fig:pipeline}
\end{figure}

\begin{figure*}[htp]
\centering
\includegraphics[width=.19\textwidth,height=.11\textheight]{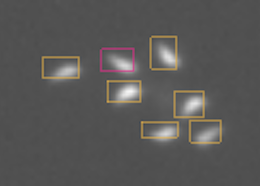}
\includegraphics[width=.19\textwidth,height=0.11\textheight]{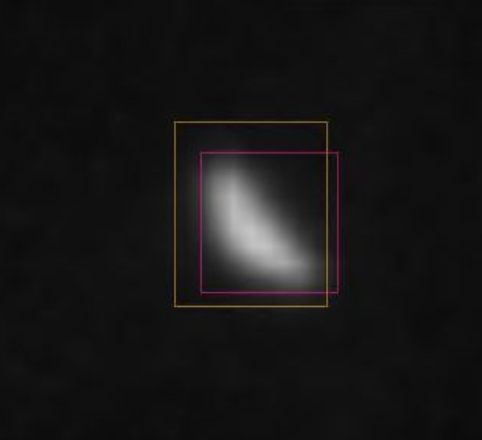}
\includegraphics[width=.19\textwidth,height=.11\textheight]{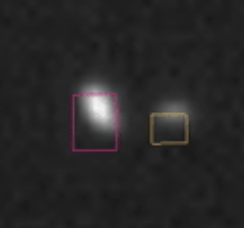}
\includegraphics[width=.19\textwidth,height=.11\textheight]{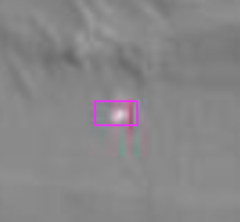}

\vspace{1pt}

\includegraphics[width=.19\textwidth,height=.11\textheight]{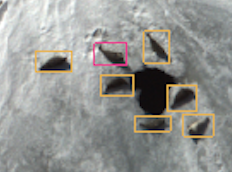}
\includegraphics[width=.19\textwidth,height=0.11\textheight]{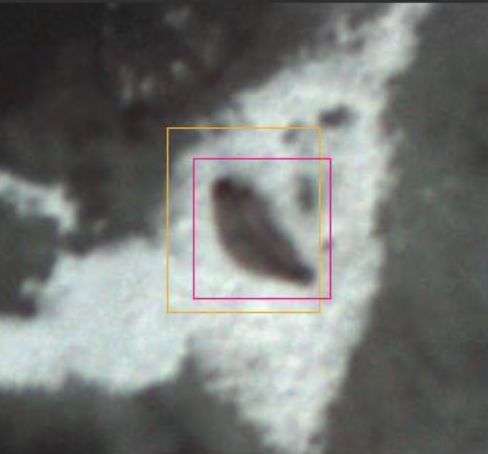}
\includegraphics[width=.19\textwidth,height=0.11\textheight]{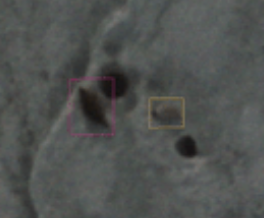}
\includegraphics[width=.19\textwidth,height=0.11\textheight]{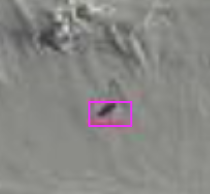}

\caption{Example detection results, IR on top, matching RGB on bottom. Purple boxes are classified as bearded seals, yellow boxes are classified as ringed seals. Some examples of double classification (column 2) or slight calibration misalignment (column 3) are shown.}
\label{fig:det_results}
\end{figure*}

These models operated in two paradigms. The first is the more traditional image-in, detections-out style of modern deep learning. Our IR hot spot detector was the standalone detector, reporting high accuracy, precision, and speed. The second is a late-fusion pipeline, illustrated in Figures \ref{fig:late_fusion} and \ref{fig:pipeline}, where the IR hot spot detector triggers the color model. The hot spots indicate likely areas of animals, and using our calibrations obtained previously, a 512x512 (polar bears at 416x416) chip is cropped out of the much larger color imagery (see Table \ref{fig:model_architecture}). These crops get passed to a second, species-specific color detector, and the results aggregated and returned. This two-stage pipeline is efficient on compute and takes full advantage of the detailed resolution of our color imagery. Example detection results are shown in Figure \ref{fig:det_results}. We implemented this late fusion in VIAME, making it reusable and ensuring reproducible results both in real time with the onboard KAMERA software and in post-processing with the VIAME GUI.

\begin{figure}[t]
  \centering
   \includegraphics[width=1\linewidth]{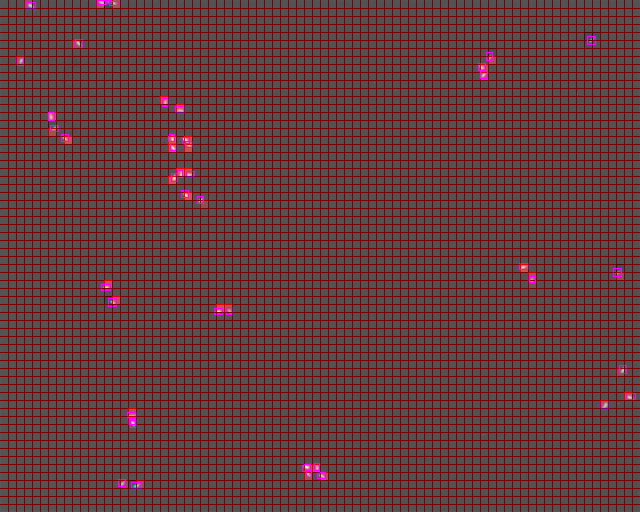}
   \caption{Example visualized feature grid $P_3/8$ representing the scale of detection for the thermal model architecture.}
   \label{fig:yolo_grid}
\end{figure}

Various model architecture modifications were experimented with based on the YOLOv3 tiny \cite{9074315} 3-layer architecture. Seal hot spots in the IR imagery are of similar size, so a final architecture with a single anchor-box detection layer was used as shown in Figure \ref{fig:yolo_grid}. The thermal dataset contained many missing labels, so the model was often penalized in training for correct guesses.  To combat this, we found that reducing the regularizer that penalizes the model in cases when no label exists ($\lambda_{noobj}$) led to a smoother training loss curve and increased validation accuracy.

For color imagery, early experiments showed suboptimal results when using the entire color image or large chips as the animals are proportionally tiny. A larger model architecture was needed to improve detection and classification accuracy, which led to inference times that exceeded our real-time requirements.  In addition, we had a large imbalance between seal classes, and an even larger imbalance with polar bear labels.  This led to a two-model approach, one color model for detecting and classifying bearded and ringed seals, and a second model for detecting only polar bears.  Due to the large class imbalance between the labels for both seal species, training used Focal Loss \cite{lin2018focallossdenseobject} to better balance the calculated error. This greatly improved classification accuracy of the seal model. Anchor boxes fitting the object size distribution and model detection layers were also chosen to maximize accuracy of both models.

As mentioned earlier, in our 2021 Beaufort survey we were able to use the IR to RGB late-fusion seal models developed since the Prosilica GT6600 color cameras were used in training and operation. During our 2025 surveys, because we had upgraded to the higher resolution Phase One color cameras and changed domains, we only used the IR hot spot detection model. The data from this latest survey is currently being used to develop models on more modern architectures including YOLOv9 \cite{wang2024yolov9learningwantlearn}. 

\subsection{Data products}

After a flight, the imagery is saved all within the same flight folder. Each image is named according to the effort name, the flight number it was collected under, its viewing angle (\underline{L}eft, \underline{R}ight, \underline{C}enter), the timestamp it was recorded, and its spectral band, i.e. \textbf{ice\_seals\_2025\_fl107\_R\_20250411\_224327.981822\_rgb}. The timestamp for the image is embedded from the INS event, so each image can be directly associated to its synchronized sample. In addition to an image for each channel, there is an associated metadata file in JSON format that contains the various parameters for each camera when the image was captured, the INS pose, the GPS location, the associated DAQ event, and human-readable text on effort name, flight, and project.

In a parallel folder there are the detection results, containing a list of every image that the object detectors ran on and a CSV file with the list of each detection that was found. There is a raw INS data file that contains the flight path, a text file containing a log of each GUI action taken, and a system configuration file, which holds the VIAME pipeline, camera models, and camera mount angle configuration.

If the user runs the \textbf{Create Flight Summary} script, they will get a single shapefile for each camera, which can be used to display the area covered from the aerial transect. The user can also run a \textbf{Detection Summary} script, which will run frame-to-frame geolocation-based tracking on all the detections, and return a single unified footprint for each camera displaying the detections overlaid onto the ground plane. These data products enable the user to quickly verify what was collected during the flight for future flight planning and survey decision making.

\begin{table*}[t]
\small
  \centering
      \begin{tabular}{@{}lccccc@{}}
        \toprule
        \multirow{2}{*}{Metric} & \multirow{2}{*}{\begin{tabular}[c]{@{}c@{}}IR Model\\(Hot Spot)\end{tabular}} & \multicolumn{3}{c}{Seal Model} & \multirow{2}{*}{\begin{tabular}[c]{@{}c@{}}Polar Bear\\Model\end{tabular}} \\
        \cmidrule(l){3-5}
        & & Overall & Ringed Seal & Bearded Seal & \\
        \midrule
        Input Dimensions & 640x512x1 & -- & 512x512x3 & 512x512x3 & 412x412x3 \\
        Benchmark Speed (FPS) & 303.7 & 231 & -- & -- & 271.2 \\
        \midrule
        True Positives (TP) & 3152 & 2928 & 2645 & 283 & 78 \\
        False Positives (FP) & 431 & 564 & 423 & 141 & 6 \\
        False Negatives (FN) & 232 & 210 & 200 & 10 & 13 \\
        \midrule
        Recall & 0.93 & 0.93 & 0.93 & 0.96 & 0.85 \\
        Precision & 0.88 & 0.84 & 0.87 & 0.67 & 0.93 \\
        F1 Score & 0.90 & 0.88 & 0.89 & 0.78 & 0.89 \\
        \bottomrule
      \end{tabular}
      \caption{Model validation results for the IR hot spot detection model, the two seal species classification models, and the polar bear detection model.  All results were evaluated on an NVIDIA GeForce GTX 1080 Ti with a batch size of 1.}
      \label{fig:ml_eval_results}
\end{table*}

\subsection{Graphical user interface}

\begin{figure}[t]
  \centering
   \includegraphics[width=1\linewidth]{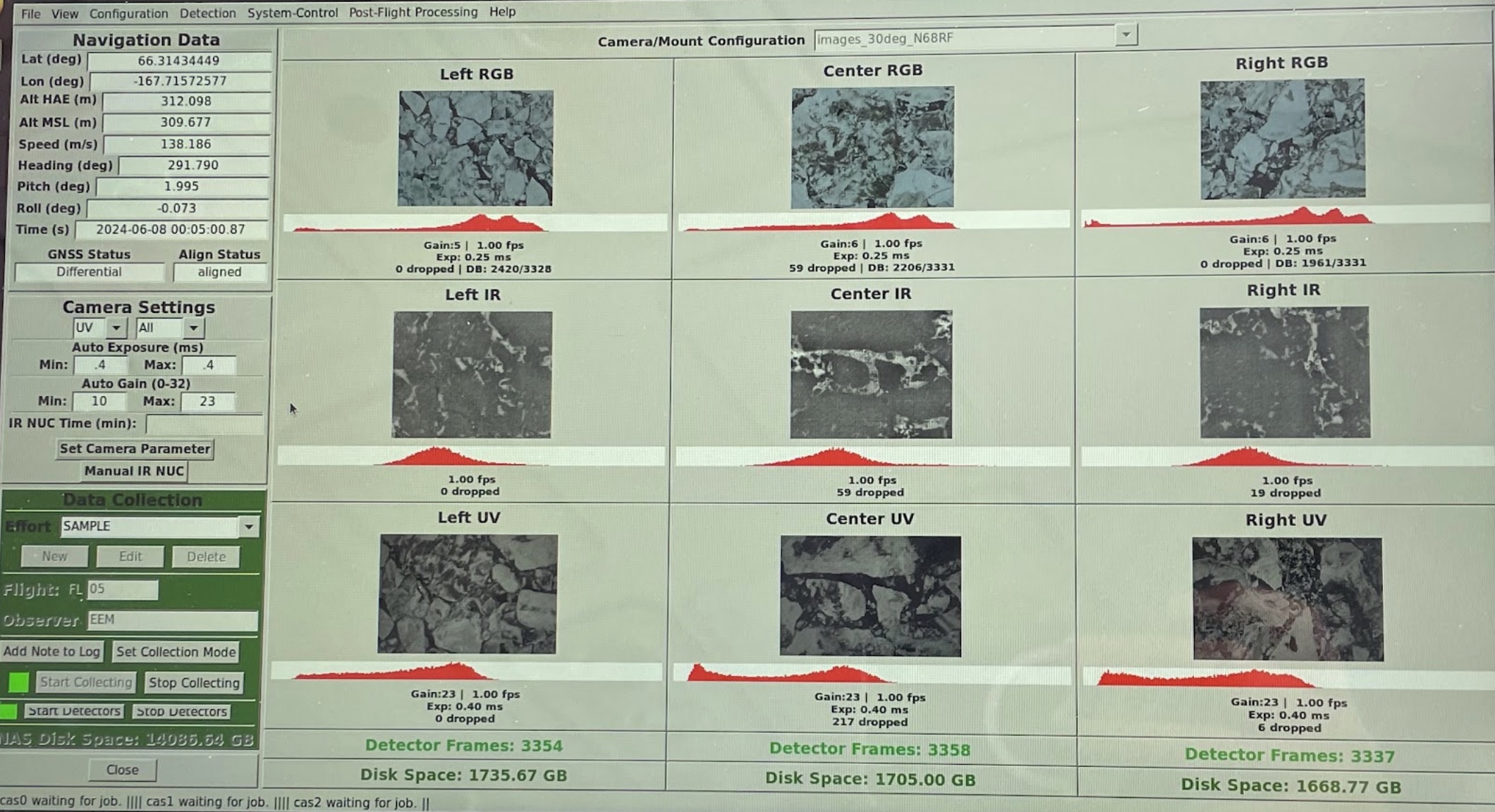}
   \caption{GUI in flight operation. All cameras are shown, with real-time options for camera control, monitoring, and collection.}
   \label{fig:gui}
\end{figure}

Because KAMERA is designed to be operated and controlled in real-time, we developed a front-end to the system (Figure \ref{fig:gui}). We created the Graphical User Interface (GUI) in WxPython, and the backend is a combination of a Redis database and JSON. This GUI has a few main goals:

\begin{enumerate}
    \item Display to the user the current imagery from all nine cameras, the current INS readings, and various system analytics such as the remaining disk space, number of frames collected and detected, and the current collecting mode.
    \item Enable the user to make on-the-fly adjustments to gain and exposure for the color / UV cameras, Non Uniformity Correction (NUC) timing for the thermal cameras, and receive real-time feedback in the form of a histogram.
    \item Give the user the ability to define the VIAME pipelines to use, as well as the camera mount configuration currently installed (e.g., 30 degrees angle).
    \item Display runtime statistics, including the number of frames dropped or if a camera has stopped streaming. This enables the user to be an active part of the system, verifying cables and hardware connections.
    \item Post-flight, give the user the option to run scripts to generate a flight summary, footprints of the area covered, or a summary of the detections.
\end{enumerate}

\noindent During the three seas survey, the GUI proved invaluable, with 12 new non-technical users operating the system.

\subsection{Open source}
In alignment with the scientific communities’ values on open science, all of the software for KAMERA is licensed under the Apache License, Version 2.0, and freely available on GitHub at \cite{kitware2025kamera_v0_4_0}. All hardware and schematics are also open. All pipefiles and object detection models are accessible under CC BY 4.0 at \cite{boss_2021_5765673}. All data will be organized and released into the public as the largest concurrent dataset of ice-associated seals and their habitat surrounding the Bering, Beaufort, and Chukchi seas ever collected.
\section{Model evaluation}
\label{sec:modeleval}
For validation, 10\% of the human labeled data was held out. Results for all three individual models are shared in Table \ref{fig:ml_eval_results}, computed on a single NVIDIA GeForce RTX 1080Ti GPU with an inference batch size of one. Our bearded seal, ringed seal, and IR hot spot detector met or exceeded a recall of 0.93, the most critical metric for us to determine data collection. Precision varied across species, with the bearded seal performing the worst at 0.67. Limited annotation data was available for polar bears, but our detector reported good recall and high precision on the data that was available. The FPS for all models was well within the limits of real-time.

In preliminary assessments of the 2025 survey, we found our IR hot spot model performed worse than the results shown here. While our recall was high enough to use for data-driven collection decisions, our precision dropped dramatically. These results show weak generalization of the IR YOLOv3 hot spot model. We attribute the drop in performance to new thermal cameras and their updated NUC calibrations. Even with identical hardware, the imagery is visually different. More analysis will be done to improve the robustness of this model in real-world conditions.
\section{Conclusion}

We have introduced KAMERA: a comprehensive system for multi-camera, multi-spectral synchronization and real-time image analysis. It has been utilized in two seal surveys, one in the Southern Beaufort, and one in the largest concurrent survey ever conducted across the Bering, Beaufort, and Chukchi seas. It improves upon previous survey methods by offering multi-camera synchronization with robust image capture, real-time detection, and precise mapping capabilities, all in an open-source software stack. 

We have illustrated an effective use of object detection for a real-world application and the difficulties that lie therein. From changing cameras to updating thermal calibrations, training and testing detection models in the environment they will be used in is critical.

The success of KAMERA depended strongly on collaboration between numerous stakeholders, including scientists, biologists, statisticians, aircraft mechanics, pilots, local airports, and hunting communities. We hope the data obtained from these surveys will continue to inform the scientific community and greater public about ice-associated seals and marine mammals as their environment changes. Since KAMERA is fully open-source and broadly applicable to wildlife surveys of many kinds, we hope the community will be able to learn from and use pieces of this software system and our methods of hardware integration.
\section*{Acknowledgements}

This research was funded by the National Oceanic and Atmospheric Administration (NOAA) through the Marine Mammal Lab (MML) under the Alaska Fisheries Science Center (AFSC), contract number 1305M323PNFFS0741. We are grateful to the original development team - Matt Brown and Michael McDermott - for their early vision, software architecture, and implementation that laid the groundwork for the current system. We also thank the NOAA flight crews, aircraft mechanics and integration specialists, scientific team, and travel coordinators that made these surveys possible. The findings and conclusions in the paper are those of the authors and do not necessarily represent the views of NOAA or the United States government.
{
    \small
    \bibliographystyle{ieeenat_fullname}
    \bibliography{main}
}

\end{document}